\pdfoutput=1

\documentclass{article}

\usepackage[final, nonatbib]{nips_2016}

\usepackage[utf8]{inputenc} 
\usepackage[T1]{fontenc}    
\usepackage{url}            
\usepackage{booktabs}       
\usepackage{amsfonts}       
\usepackage{amsmath}
\usepackage{nicefrac}       
\usepackage{microtype}      
\usepackage{graphicx}
\usepackage{algorithm}
\usepackage[noend]{algorithmic}
\usepackage{multirow}
\usepackage{array}
\usepackage{caption} 
\usepackage[
bookmarks=true,
colorlinks=true,
linkcolor=red,
urlcolor=blue,
citecolor=blue,
pdftex,
bookmarks=true,
linktocpage=true, 
hyperindex=true
]{hyperref}

\usepackage{hyperref}       

\hypersetup{
backref=true
bookmarksnumbered,
pdfstartview={FitH},
citecolor={blue},
linkcolor={red},
urlcolor={black},
pdfpagemode={UseOutlines}
}

\usepackage{capt-of}

{%
\newcommand\T{\rule{0pt}{4.5ex}}       
\newcommand\B{\rule[-3.5ex]{0pt}{0pt}} 

\title{Deep neural networks are robust to weight binarization and other non-linear distortions}

\author{
  \hspace*{-0.3cm} Paul Merolla, 
  Rathinakumar Appuswamy, 
  John Arthur, 
  Steve K. Esser, 
  Dharmendra Modha  \\
  IBM Research--Almaden \\
  \texttt{\{pameroll,rappusw,arthurjo,sesser,dmodha\}@us.ibm.com} 
}

\begin{document}

\maketitle

\begin{abstract}
Recent results show that deep neural networks achieve excellent performance even when, during training, weights are quantized and projected to a binary representation.
Here, we show that this is just the tip of the iceberg: these same networks, during testing, also exhibit a remarkable robustness to distortions beyond quantization, including additive and multiplicative noise, and a class of non-linear projections where binarization is just a special case.
To quantify this robustness, we show that one such network achieves $11\%$ test error on CIFAR-10 even with $0.68$ effective bits per weight.  
Furthermore, we find that a common training heuristic---namely, projecting quantized weights during backpropagation---can be altered (or even removed) and networks still achieve a base level of robustness during testing.
Specifically, training with weight projections other than quantization also works, as does simply clipping the weights, both of which have never been reported before.
We confirm our results for CIFAR-10 and ImageNet datasets. 
Finally, drawing from these ideas, we propose a stochastic projection rule that leads to a new state of the art network with $7.64\%$ test error on CIFAR-10 using no data augmentation.
\end{abstract}

\section{Introduction}

Deep neural networks (DNNs) trained using backpropagation have been shown to perform exceptionally well on a wide range of classification tasks~\cite{krizhevsky2012imagenet, silver2016mastering, lecun2015deep, schmidhuber2015deep, johnson2015densecap}.  Typically, these networks use a high precision representation for weights (e.g., 32-bit floating point) for both training and for inference.  Considering just inference tasks, a long standing goal has been to reduce the precision of weights without sacrificing performance, with the aim of lowering the network's computational and memory footprint.  This has practical applications that include network compression, running networks faster and more efficiently on conventional hardware~\cite{han2015learning,rastegari2016xnor}, and running networks on specialized hardware designed specifically for reduced precision representations~\cite{merolla2014million, esser2016convolutional}.

Remarkably, a slew of recent work has shown that using just two (binary) or three (ternary) values for weights, DNNs can approach state of the art performance on popular benchmarks~\cite{courbariaux2015binaryconnect,esser2016convolutional,rastegari2016xnor, esser2015backpropagation}.  The basic approach is to apply gradient descent using high precision weights during training, and {\it project} the weights via a quantization function (such as the $\mathrm{sign}$ function) in forward/backward passes.  This way, the high precision weights are able to accumulate small gradient updates---computed with respect to the projected ones---allowing the network to explore discrete configurations in a continuous setting.  It has been argued, rather reasonably, that this training procedure is essential to learning low precision representations~\cite{courbariaux2015binaryconnect}.  In this work, we suggest a more nuanced view.  We shift the focus from the idea that projecting quantized weights during training leads to networks robust to that quantization, to a more general phenomenon, where projecting weights via certain functions (not necessarily quantization) lead to networks robust not only to that function (or distortion), but to an entire class of distortions.  This is as if a patient given the flu vaccine, finds themselves inoculated against measles, mumps, and malaria as well.

Here, we present a number of new findings:
\begin{enumerate}
\item{We show that many networks that perform well when their weights are binarized, also perform well for other kinds of distortions.  These distortions include additive and multiplicative noise, as well as applying non-linear distortions to the weights.}

\item{We report that using weight projections other than quantization during training also lead to robust networks. Furthermore, we show examples where standard backprop with weight clipping can learn a base level of robustness, although performance is slightly reduced.} 

\item{Based on these observations, we propose a new stochastic projection rule inspired by BinaryConnect~\cite{courbariaux2015binaryconnect}, except our rule projects weights to random intervals (as opposed to quantized values). This rule results in new state of the art performance for CIFAR-10 with $8.25\%$ in the binary case and $7.64\%$ in the non-binary case.}
\end{enumerate}

The organization of our paper is as follows: After reviewing related work in Section~\ref{sec:rw}, we describe our training algorithm in Section~\ref{sec:dnn-pw}.  Next in Section~\ref{sec:results}, we use this algorithm to train six networks on the CIFAR-10 dataset, exploring different combinations of weight projections and weight clipping parameters.  To help frame our results, we first delve into a curious finding in Section~\ref{subsubsec:hp1}.  Specifically, we show that a network trained using quantized weights also has low test error for non-quantized weights, suggesting that it is robust to weight distortions.  Following this lead in Section~\ref{subsec:r-w-proj}, we uncover that this network (and others) are robust to distortions beyond weight quantization.  In Section~\ref{subsubsec:learn} we try to tease apart the aspects of training that lead to these robust networks, and report for the first time that non-quantized projections (and even no projections at all) can also lead to robustness.  A new stochastic projection rule is explored in Section~\ref{subsec:smp}.  Then in Sections~\ref{subsec:imagenet} and~\ref{subsec:image-bench} we check that our findings hold up for ImageNet.  Section~\ref{sec:BP-w-proj} puts forth two theoretical ideas on how backprop is able to find these robust solutions.  Finally, we conclude and discuss future directions in Section~\ref{sec:concl}.

\section{Related work}
\label{sec:rw}

Our current work is related to a flurry of recent research on DNNs with low precision weights.  The major advance has been the discovery of a backprop-based learning method that quantizes weights during propagations.  This training method was introduced in~\cite{courbariaux2014training, courbariaux2015binaryconnect} with impressive results on CIFAR-10, and developed in the context of neuromorphic hardware in~\cite{esser2016convolutional}.  \cite{stromatias2015robustness} proposed a similar rule for contrastive divergence.  Here we build on this work by exploring a more general class of weight projections that are not restricted to quantization.  Another recent approach approximates weights using binary values during training, formulated as a constrained optimization problem, with promising results on ImageNet~\cite{rastegari2016xnor}.  We expect their findings are consistent with our results, however, this is left for future work.  Other approaches have developed a probabilistic interpretation of discrete weights~\cite{soudry2014expectation, esser2015backpropagation}, however, they have not yet been extended to convnets or datasets larger than MNIST.

\section{DNNs with projected weights}
\label{sec:dnn-pw}

We consider a DNN as two sets of tensors: input activations $A$, and weights $W$, where each set is indexed by its layer $k$.  The output of a layer (which is also the input to the next layer) is computed using the convolution operation $*$, for example $A_{k+1} = r(A_k * \mathrm{Proj}(W_k))$ where $r$ is the typical ReLU activation, and $\mathrm{Proj}$ is a projection operator described below.  For notational simplicity, we specify fully-connected layers as convolutions, and do not consider neuron biases.  

In this work, we explore DNN's under various projections to the weights used for training and testing.  These projections are defined in Table~\ref{tab:proj} as scalar functions and extended to operate on tensors element-wise.  We denote the $i$-th element of tensor $W_k$ by $w_{ki}$.  We introduce a per layer factor $\alpha_k=\underset{i}{\mathrm{max}}(|w_{ki}|)$ to normalize weights to the interval $[-1, 1]$ for certain projections.  While normalizing across the entire layer seems crude compared to a filter-wise normalization (as in ~\cite{rastegari2016xnor}), we have found that the two cases lead to similar results.  It is worth pointing out that the $\mathrm{Power}$ projection generalizes $\mathrm{Sign}$ and $\mathrm{None}$, since $\mathrm{Power}(w_{ki}, 0)=\mathrm{Sign}(w_{ki})$, and $\mathrm{Power}(w_{ki}, 1)=\mathrm{None}(w_{ki})$.

The procedure that we use to train is described in Algorithm~\ref{alg:DNN}, which is similar to BinaryConnect~\cite{courbariaux2015binaryconnect} except we allow for arbitrary weight projections.  The differences between our algorithm and standard backprop, are i) we first project the weights using $\mathrm{Proj}(W_k, \theta)$ before computing the forward pass, ii) we compute the gradients with respect to the projected weights $P$ but apply updates to $W$, and iii) we clip the weights after each update via a per layer clip value $c_k$.  $c_k$ is defined as the standard deviation of the initialized weights (from~\cite{glorot2010understanding}) scaled by a global factor $f$, where $f=0.5$ unless otherwise noted.  This algorithm reduces to standard backprop when the projection is $\mathrm{None}$ and $c_k=\infty \ \forall \ k$.  

Testing, just like training, is also performed for a particular projection, however it is important to note that testing and training projections are independently specified.  We often refer to projections applied during testing as {\it distortions}.

\begin{table}[t]
\centering
\footnotesize
\tabcolsep=0.11cm
\scalebox{0.9}{
\begin{tabular}{ | l | l | l | l |}
	\hline
	{\bf Projection} 	&	{\bf Definition} 	& {\bf Allowed states} & {\bf Use} \\
	\hline
	\hline
	$\mathrm{Sign}(w_{ki})$		&$\alpha_k \mathrm{sign}(w_{ki})$	& $ \{-\alpha_k, +\alpha_k\}$	& Test and Train \\
	\hline
	$\mathrm{Round}(w_{ki})$ 	& $\alpha_k \mathrm{round}(\frac{w_{ki}}{\alpha_k})$ 	& $\{-\alpha_k , 0, +\alpha_k\}$		& Test and Train	\\[0.5ex]
	\hline
	$\mathrm{None}(w_{ki})$ 	& $w_{ki}$ 	& $(-\infty, \infty)$  	& Test and Train \\
	\hline
	$\mathrm{Power}(w_{ki}, \beta)$ 	& $\alpha_k |\frac{w_{ki}}{\alpha_k}|^\beta \mathrm{sign}(w_{ki})$ 	&  $[-\alpha_k, +\alpha_k]$ 		& Test and Train \\[0.5ex]
	\hline	
	$\mathrm{Stoch}(w_{ki})$	&
	$
	\begin{cases}
   		+\alpha_k,	& \text{with probability } p=\frac{1}{2}(\frac{w_{ki}}{\alpha_k} + 1) \\
    		-\alpha_k,    & \text{with probability } 1-p
	\end{cases}
	$ 	& $\{-\alpha_k, +\alpha_k\}$	& Train
	\T \B \\  
	\hline
	$\mathrm{StochM}(w_{ki},\gamma)$	& 
	$
	\begin{cases}
   		+w_{ki} \mathrm{U}(\gamma,\frac{1}{\gamma}) ,	& \text{with probability } p=\frac{1}{2}(\frac{w_{ki}}{\alpha_k} + 1) \\
    		-w_{ki} \mathrm{U}(\gamma,\frac{1}{\gamma}) ,	& \text{with probability } 1-p
	\end{cases}
	$	& $[-\frac{\alpha_k}{\gamma}, +\frac{\alpha_k}{\gamma}]$	& Train	 
	\T \B \\  
	\hline
	$\mathrm{AddNorm}(w_{ki},\sigma)$	& $w_{ki} + \mathrm{N}(0,\alpha_k \sigma)$ 	& $(-\infty, \infty)$  	& Test \\
	\hline
	$\mathrm{MultUnif}(w_{ki},\gamma)$	&  $w_{ki} \times \mathrm{U}(\gamma,\frac{1}{\gamma})$ 	& $ [-\frac{\alpha_k}{\gamma}, +\frac{\alpha_k}{\gamma}]$ 	& Test \\[0.5ex]
	\hline
\end{tabular}
}
\\
\caption {Definitions for projections. $\mathrm{N}(0,\sigma)$ is drawn from a normal distribution with mean $0$ and standard deviation $\sigma$, and $\mathrm{U}(a,b)$ is drawn from a uniform distribution in the range $[a, b]$.}
\label{tab:proj}
\end{table}

\renewcommand{\algorithmicrequire}{\textbf{Input:}}
\renewcommand{\algorithmicensure}{\textbf{Output:}}
\begin{algorithm}[thb]
{\footnotesize
  \caption{Training a DNN for a weight projection.  $\mathrm{Proj}(W_k, \theta)$ is a projection from Table~\ref{tab:proj} with parameter $\theta$, $c_k$ are clip boundaries, $L$ is the loss function, and $N$ is the number of layers.}
  \label{alg:DNN}
  \begin{algorithmic}[1]
  \REQUIRE Minibatch of inputs $I$ and targets $T$, current weights $W^t$, and learning rate $\lambda$.   
  \ENSURE Updated weights $W^{t+1}$. 
    \STATE {\bf Project weights:}
    \FOR{$k=1$ to $N$}
      \STATE $P_k = \mathrm{Proj}(W_k^t, \theta)$~~//~{\scriptsize In standard backprop, $\mathrm{Proj}$ is $\mathrm{None}$, so $P_k=W_k^t$.}
    \ENDFOR
    \STATE {\bf Forward propagation:}
    \STATE $O = \mathrm{Forward}(I,P)$~~//~{\scriptsize Standard forward pass, computed with respect to projected weights $P$.}
    \STATE {\bf Backward propagation:}
    \STATE $\frac{\partial L}{\partial P} = \mathrm{Backward}(\frac{\partial L}{\partial O}, P)$~~//~{\scriptsize Standard backward pass, also computed with respect to $P$.}
    \STATE {\bf Update weights:}
    \STATE $\hat{W} = \mathrm{Update}(W^t, \frac{\partial L}{\partial P}, \lambda)$~~//~{\scriptsize Updates applied to $W^t$, can use any update rule (e.g., SGD, ADAM, etc).}
    \STATE {\bf Clip weights:}
    \FOR{$k=1$ to $N$}
      \STATE $W^{t+1}_k = \mathrm{max}(\mathrm{min}(\hat{W_k},c_k), -c_k)$~~//~{\scriptsize Weights remain in range $[-c_k, c_k]$. In standard backprop, $c_k=\infty$ .}  
    \ENDFOR    
  \end{algorithmic}
  } \normalsize
\end{algorithm}

\section{Results}
\label{sec:results}

In this section, we explore the performance and robustness of networks trained using different weight projections and clip settings, on both CIFAR-10 (Section~\ref{subsec:cifar}) and ImageNet (Section~\ref{subsec:imagenet}).   

\subsection{CIFAR-10 experiments}
\label{subsec:cifar}

CIFAR-10 is an image classification dataset of small ($32 \times 32$) color images with $10$ categories~\cite{krizhevsky2009learning}.  For our experiments, we do not use data augmentation.  To train, we use the ADAM learning rule~\cite{kingma2014adam} with a learning rate of $0.003$, and a batch size of $50$; a square hinge loss; and batch normalization~\cite{ioffe2015batch}.  These results were obtained in TensorFlow~\cite{abadi2016tensorflow}, except for a control network that used Caffe~\cite{jia2014caffe}.

Our experimental flow is as follows: Using the training set pre-processed with global contrast normalization and ZCA whitening from Pylearn2~\cite{pylearn2_arxiv_2013}, we trained six networks for 500 epochs using a DNN with 6 Conv and 2 FC layers (Figure~\ref{fig:two_case}A).  These networks are named according to their training parameters.  For example Tr-Sign-C specifies a network trained (Tr) using the $\mathrm{Sign}$ projection with clipping (C); we append NC for no clipping when $c_k=\infty$.  A seventh network (NiN-ctrl) was downloaded pre-trained from the Caffe model zoo~\cite{lin2013network} for comparison.  After training we evaluate each network's test error for different distortions to the weights (and no distortions to the biases).  Batch norm parameters are re-computed on the training data for each distortion.  Tests are specified using a similar naming convention, for example Te-Power refers to a test (Te) for the $\mathrm{Power}$ projection.  Results for Te-None, Te-Sign, and Te-Round are summarized in Table~\ref{tab:cifar10} along with comparisons to prior work.  There are a few surprising results in this table, and we analyze them one at a time in the next sections.

\begin{table}[h]
\centering
\footnotesize
\tabcolsep=0.11cm
\scalebox{0.9}{
\begin{tabular}{l l l l l l}
	\hline
	Network			& Te-None 	& Te-Sign				& Te-Round 		& Notes 	\\
	\hline
	Tr-None-NC		 & $11.35\%$ 		& $14.92\%$		& $90.0\%$ 		& \\	
	Tr-None-C			 & $9.97\%$ 		& $11.32\%$		& $11.2\%$ 		& \\
	Tr-Sign-C			 & $10.64\%$ 		& $9.95\%$		& $10.4\%$ 		& \\
	Tr-Stoch-C			 & $8.12\%$ 		& $8.38\%$		& $8.36\%$ 		& \\
	Tr-Power-C		 & $10.3\%$ 		& $10.6\%$		& $10.2\%$ 		&  $\beta \in \mathrm{U}[0,2]$ \\
	Tr-StochM-C		 & ${\bf 7.64\%}$ 	& ${\bf 8.25\%}$	& $7.88\%$ 		& $\gamma=0.5$  	\\
	\hline
	NiN-ctrl~\cite{lin2013network}							& $10.4\%$ 	& 				&  \\
	BC-Sign~\cite{courbariaux2015binaryconnect}		& 	 			& $9.90\%$	&  \\
	BC-Stoch~\cite{courbariaux2015binaryconnect}		& $8.27\%$ 	& 				&  \\		
	\hline
\end{tabular}
}
\\
\caption {Test error on CIFAR-10.  Rows are networks trained using different weight projections and clipping parameters, and bottom three rows are from the literature; columns are tests using different weight distortions.}
\label{tab:cifar10}
\end{table}

\subsubsection{High precision or -1,+1?  It doesn't (much) matter}
\label{subsubsec:hp1}

\begin{figure}[t]
\begin{tabular}{ll}
\multicolumn{2}{c}{
\includegraphics[scale=1.0]{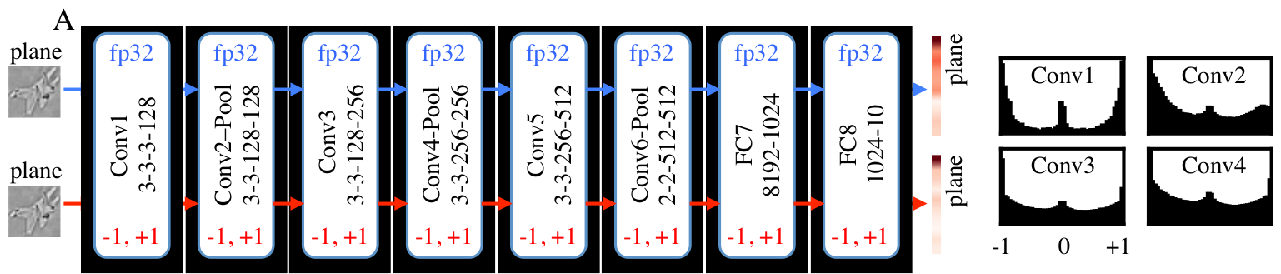}
}
\\
\\
\multirow{2}{*}{
\includegraphics[trim=0 0 0 50]{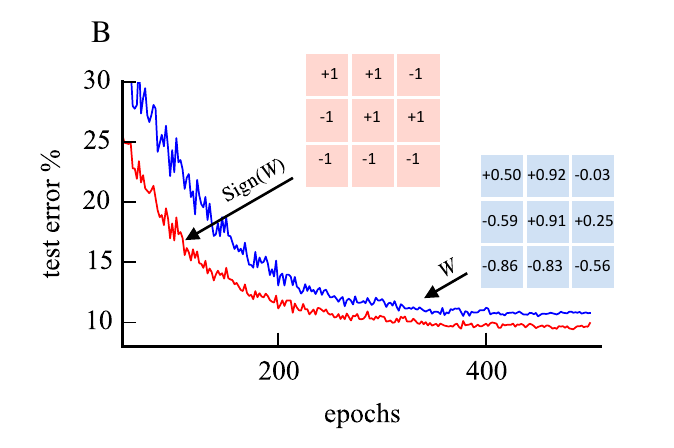}
}
&
\includegraphics[scale=1.0]{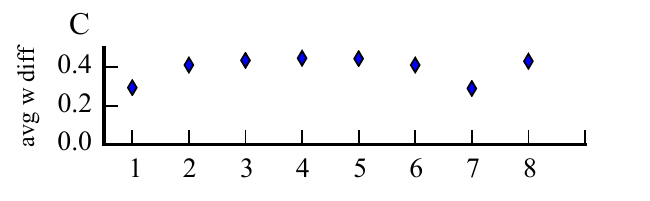} 
\\
&
\includegraphics[scale=1.0]{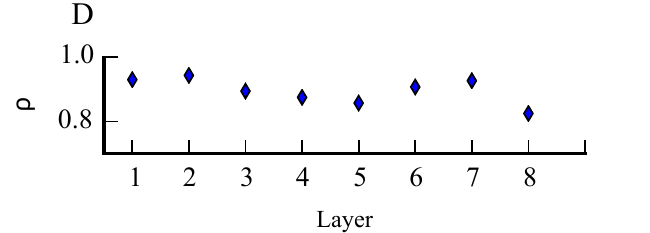}
\\
\end{tabular}
\caption{
  (A) CIFAR-10 network trained using $\mathrm{Sign}$ projection (Tr-Sign-C) where each Conv is specified as $x$-$y$-$n_{in}$-$n_{out}$, and FC as $n_{in}$-$n_{out}$.  There are two test scenarios, Te-None (top half with fp32), and Te-Sign (bottom half with binary).  Four weight histograms post training are shown on the right.
  (B) Test error during training, evaluated every two epochs for Te-None and Te-Sign.  Insets show weights for two corresponding filters post training.
  (C) Average absolute differences between $W_k$ and $\mathrm{Sign}(W_k)$ at each layer.
  (D) Correlation coefficient between neuron activity at each layer for a minibatch during two forward passes, one evaluated using $W_k$ and the other using $\mathrm{Sign}(W_k)$.
  }
\label{fig:two_case}
\end{figure}

One interesting result is that all of our networks have comparable test errors for Te-None and Te-Sign.  To see why this is surprising, consider the DNN from Figure~\ref{fig:two_case}A trained using the $\mathrm{Sign}$ projection (Tr-Sign-C).  During training, the loss is computed with respect to the sign of the weights, and not the high precision weights directly.  Yet this network performs well when it is evaluated using {\it either} binarized weights or the high precision ones (Figure~\ref{fig:two_case}B).  This result would be expected if the two weight distributions converged to the same values, however, this is not the case: For example, the weights of two corresponding filters are noticeably different (Figure~\ref{fig:two_case}B, insets), and these quantization errors are present throughout all the layers (Figure~\ref{fig:two_case}C).  Yet despite these differences, the activity in the network still converges to similar patterns as it propagates through the layers (Figure~\ref{fig:two_case}D), demonstrating a surprising insensitivity to the exact weight values.  Based on these observations, we next explore if these networks are also robust to other non-linear distortions beyond weight binarization.

\subsubsection{Robustness to weight distortions beyond -1,+1}
\label{subsec:r-w-proj}

We investigate the premise that networks that perform well with binary weights, also perform well for many types weight distortions.  Here, we focus on networks trained using quantization-based weight projections (based on the networks from BinaryConnect), where allowed weight states during training are discrete.  Specifically, we consider Tr-Sign-C and Tr-Stoch-C under three distortions: Te-AddNorm, Te-MultUnif, and Te-Power.  NiN-ctrl is also shown for comparison.

\begin{figure}[t]
\begin{tabular}{ll}
\includegraphics[scale=1.0]{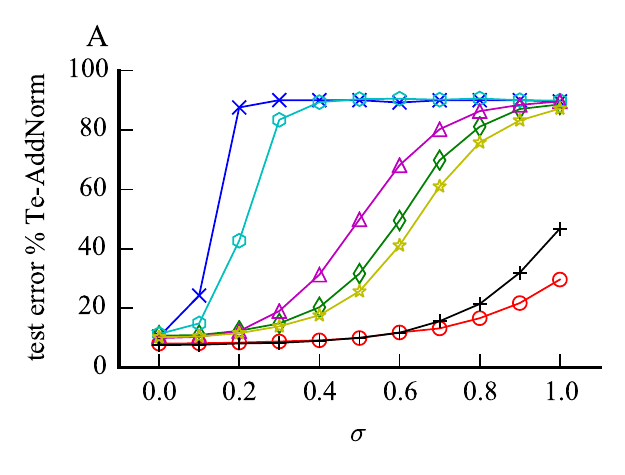}
&
\includegraphics[scale=1.0]{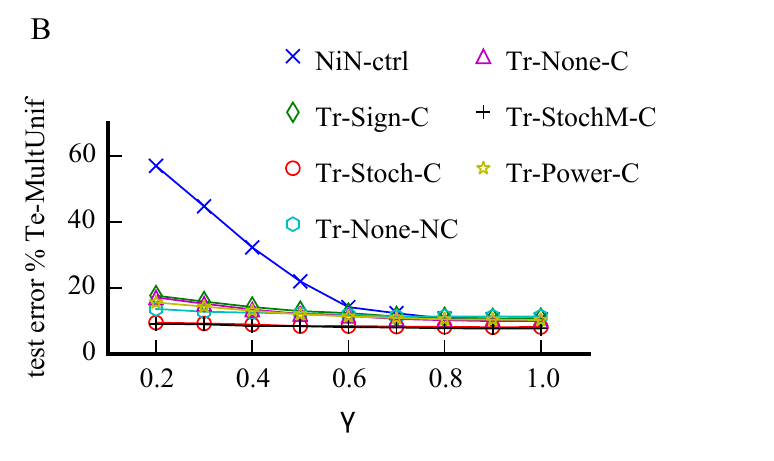}
\\
\includegraphics[scale=1.0]{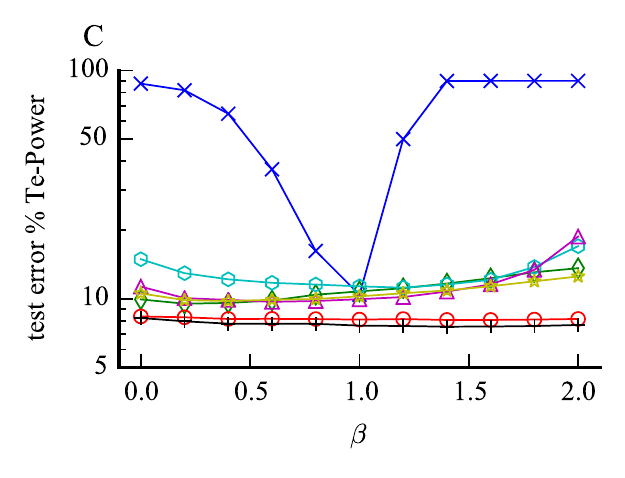}
&
\includegraphics[scale=1.0]{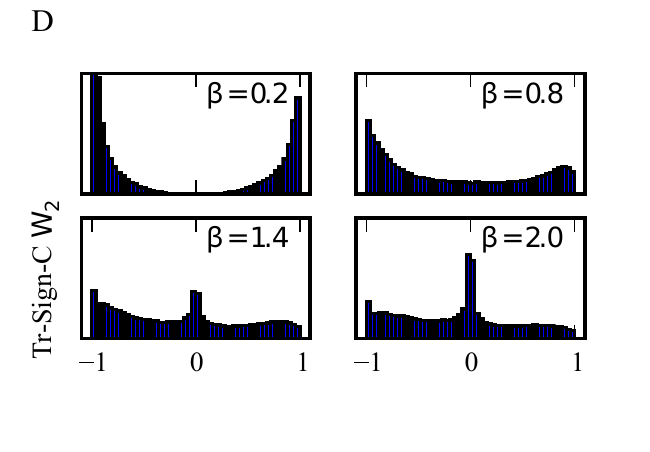}
\end{tabular}
\caption{  
  {\bf DNNs are robust to different types of weight distortions.}  Six networks were trained using different projections and clip values.  After training, each network (including a control network) was tested with gaussian (A) and multiplicative (B) noise applied to the weights, and a distortion where each weight is raised to a power (C).  Weight histograms for Conv2 of Tr-Sign-C are shown, where weights are projected using $\mathrm{Power}$ for four values of $\beta$ (D).   
  }
\label{fig:robust_noise}
\end{figure}

In the case of adding Gaussian noise to each weight (Te-AddNorm), we observe that test error increases with $\sigma$ (Figure~\ref{fig:robust_noise}A), however, Tr-Sign-C and Tr-Stoch-C are significantly more resilient than NiN-ctrl.  In particular Tr-Stoch-C achieves a test error of $11\%$ even when $\sigma=0.55$.  This corresponds to $0.68$ bits per weight using a signal to noise analysis: $\frac{1}{2}\mathrm{log}_2(1 + \frac{Q_w}{Q_n})$, where $Q_w$ and $Q_n$ are the second moments of the weight and noise distributions respectively.  In the case of multiplicative noise applied to each weight (Te-MultUnif), the trend is similar where Tr-Stoch-C and Tr-Sign-C are more resilient to noise (smaller $\gamma \rightarrow$ higher noise) compared to the control (Figure~\ref{fig:robust_noise}B). 

Finally for Te-Power, each weight is normalized to $[0,1]$, raised to the power $\beta \in [0, 2]$, and multiplied by its sign.  Remarkably, Tr-Stoch-C and Tr-Sign-C are also robust to these types of distortions.  This is surprising because these networks were never explicitly trained using them (Figure~\ref{fig:robust_noise}C, note the semi-log scale).  To visualize how this projection distorts the weights for different values of $\beta$, we show Tr-Sign-C's weight distribution for Conv2 (Figure~\ref{fig:robust_noise}D); note that lower $\beta$ pushes the weights to a bi-modal distribution that approaches $\mathrm{Sign}(W_k)$.  In contrast, NiN-ctrl is sensitive to these distortions, and only has low error for low distortion with $\beta$ near $1$. 

Bringing these results together, it appears that backprop is finding paths to solutions that are generally robust to weight distortions, and not just the specific quantizations used during training.  

\subsubsection{Learning robust networks with and without weight projections}
\label{subsubsec:learn}

In this section, we attempt to uncover the aspects of training that lead to robust networks.  Our first experiment is to see if projections other than weight quantization also work.  Accordingly, we trained a network using the $\mathrm{Power}$ projection (Tr-Power-C) where a new $\beta \in [0,2]$ is drawn for each minibatch.  The network converges to a solution that is similarly robust to Tr-Sign-C (Figure~\ref{fig:robust_noise}A-C), which is remarkable considering that the projected weights undergo a stochastic non-linear distortion at each training step.  This confirms that training with weight projections other than quantization also works, and opens the door to trying more exotic projections (for example, see Section~\ref{subsec:smp}).

Next, we tried removing weight projections altogether.  We trained networks Tr-None-C (no projection with clipping) and Tr-None-NC (no projection without clipping).  Putting these networks through the same battery of tests, we observe that Tr-None-C is more robust than Tr-None-NC, although they both exhibit the same basic trends.  Notably Tr-None-C still achieves a test error of $11.3\%$ even when its weights are quantized to binary, even though it was never trained with binary weights.  Later on in Section~\ref{sec:BP-w-proj}, we hypothesize how weight clipping can be viewed as a type of regularization, which may help explain these results.

A curious result is that Tr-None-NC also exhibits a base level of robustness while NiN-ctrl does not---even though they are both trained using standard backprop without clipping.  This conflicts with previous findings where crude quantization does not lead to good performance~\cite{moerland1997neural}.  We believe that typically, the process of ``tuning'' a DNN for the best score often leads to non-robust networks.  In our case, however, our networks are first ``tuned'' using weight projections (e.g., $\mathrm{Sign}$) during training.  Therefore, when we remove the weight projections from training, the rest of the parameters are still in a regime where backprop discovers robust solutions.  Because this learning regime appears delicate, we take a more practical view in the rest of the paper and focus on backprop with weight clipping and (or) weight projections.

\subsubsection{Stochastic multiplicative projection}
\label{subsec:smp}

Inspired by the $\mathrm{Stoch}$ projection first introduced in~\cite{courbariaux2015binaryconnect}, we constructed a new stochastic projection rule, $\mathrm{StochM}$.  In $\mathrm{Stoch}$, each weight is randomly projected to $+1$ with probability $p=\frac{1}{2}(\frac{w_{ki}}{\alpha_k} + 1)$ and $-1$ with probability $1-p$.  $\mathrm{StochM}$ derives from a similar idea, but now projects each weight to the interval $[\gamma w_{ki}, \frac{w_{ki}}{\gamma}]$ with probability $p$, and $[-\gamma w_{ki}, -\frac{w_{ki}}{\gamma}]$ with $1-p$.  The rationale is that there is nothing special about projecting to two values, so why not sample the weight space more densely?

In terms of performance, Tr-StochM-C achieves $7.64\%$ error for Te-None and $8.25\%$ for Te-Sign (Table~\ref{tab:cifar10}), which are state of the art for this data set without data augmentation, to the best of our knowledge.  The network also exhibits a high degree of robustness (Figure~\ref{fig:robust_noise}A-C).  It is interesting to note that using $\gamma=0.5$ during training, the expected value of the projected weight is no longer the same as $w_{ki}$, which was originally thought to be important for these stochastic projections to work properly~\cite{courbariaux2015binaryconnect}.

\subsection{Towards a robust AlexNet}
\label{subsec:imagenet}

To see whether our results extend beyond CIFAR-10, we moved to ImageNet (ILSVRC2012), which is a dataset with $\sim1.2$M training images and 1K classes~\cite{ILSVRC15}.  We use an AlexNet~\cite{krizhevsky2012imagenet} with $5$ convolution and $2$ fully connected layers, modified with batch norm layers.  These experiments were run in MatConvnet~\cite{vedaldi2015matconvnet} using SGD with no momentum and a batch size of 512.

Before benchmarking, we tested whether weight projections are needed to obtain robust networks.  Accordingly, two AlexNets were trained without projections for $20$K iterations (8.5 epochs): one without weight clipping (Tr-None-NC), and one with weight clipping (Tr-None-C).  In both cases the top-5 error was reported for Te-None, Te-Round, and Te-Sign.  Focusing on network Tr-None-NC (Figure~\ref{fig:imnet}A), we find that while Te-None reaches $28.1\%$\footnote{This network reaches $22.5\%$ after $100$K iterations.}, the test error for Te-Round and Te-Sign are significantly worse.  So our previous CIFAR-10 result for Tr-None-NC did not hold up for ImageNet. 
 
Network Tr-None-C tells a different story (Figure~\ref{fig:imnet}B): the network has similar performance for Te-None, Te-Round, and Te-Sign, at $35\%$, $34.7\%$ and $40.5\%$ respectively.  This confirms our previous observation that weight clipping during training can influence network robustness.  Although in this case, the peak error after $20$K iterations is about $7\%$ higher for Te-None compared to network Tr-None-NC; to help to mitigate this effect, we use a clipping scheduling to increase clip values during training in later experiments.  Also we observe that, Te-Round performs better than Te-Sign.

\begin{figure}[t]
\begin{tabular}{ll}
\includegraphics[scale=1.0]{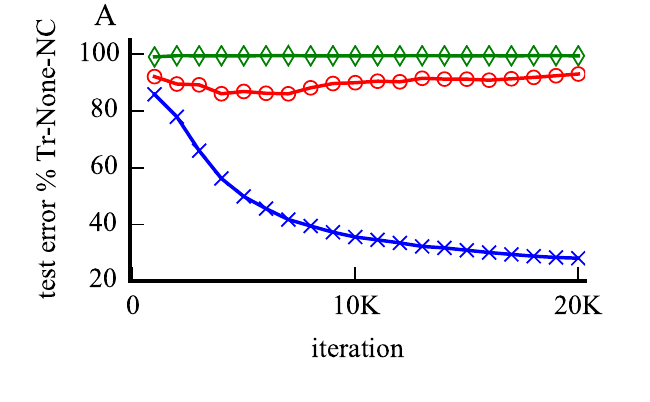}
&
\includegraphics[scale=1.0]{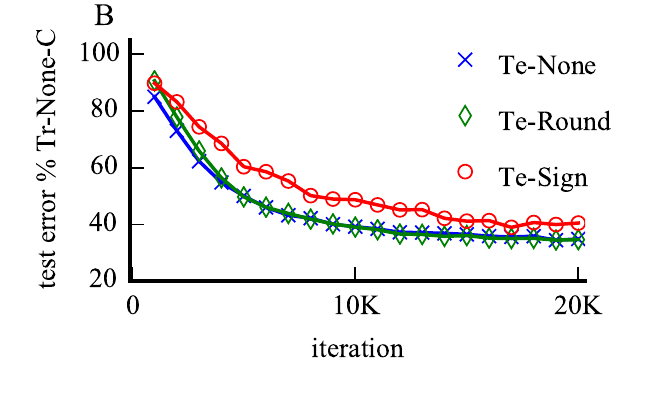}
\end{tabular}
\caption {Two versions of AlexNet were trained without weight projections, Tr-None-NC which does not use weight clipping (A) and Tr-None-C which uses weight clipping (B).  Test error was computed every $1$K iterations for projections $\mathrm{None}$, $\mathrm{Round}$, and $\mathrm{Sign}$.  With weight clipping alone, the network becomes robust to weight quantizations.
  }
\label{fig:imnet}
\end{figure}

\subsubsection{ImageNet benchmarks}
\label{subsec:image-bench}

We benchmarked Tr-StochM3-C\footnote{$\mathrm{StochM3}$ is the three-value counterpart to $\mathrm{StochM}$.  Specifically, the middle $50\%$ of the weights are dropped to $0$ with probability $0.5$.} on ImageNet.  This network was trained for $150$K iterations (64 epochs).  The learning rate was dropped from $0.1$ by $0.01 \times$ at iterations $100$K and $125$K.  We also increased clip values by scaling global clip factor $f$ from $4.5$, by $1.4 \times$ at iterations $5$K, $10$K, and $15$K.  This had the effect of keeping Te-None and Te-Round scores in sync, while also allowing the network to reach lower error.\footnote{Note that when the clip values are scaled, performance temporarily dips as the network adapts.}  We probed the network's robustness using Te-Power, and compare test error to two recent models that use binary weights (Figure~\ref{fig:imnet}).

\begin{figure}[t]
\centering
\begin{minipage}[t]{.48\textwidth}
\centering
\vspace{-10pt}
\includegraphics[scale=1.0]{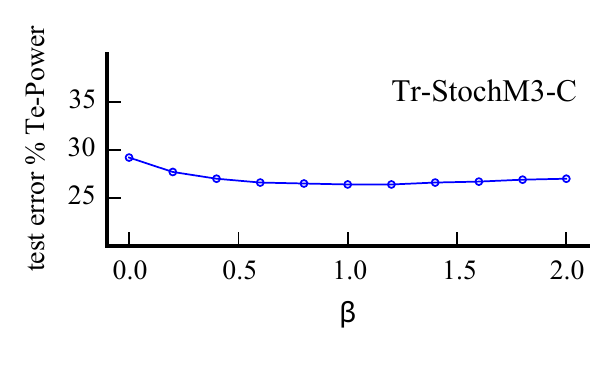}
\end{minipage}\hfill
\begin{minipage}[t]{.48\textwidth}
\centering
\vspace{10pt}
\scalebox{0.8}{
\begin{tabular}{l l l l l l}
	\hline
	Network					& Te-None 	& Te-Sign 		& Te-Round	 \\
	\hline
	Tr-StochM3-C				& $26.4\%$ 	& $29.2\%$	& $26.6\%$	 \\
	\hline
	BC-Sign~\cite{rastegari2016xnor}		& 		 		& $39\%$ 			& 			\\
	BWN~\cite{rastegari2016xnor}			&  		 	& ${\bf 23\%}$ 	& 				\\
	\hline
\end{tabular}
}
\vspace{0.45in}
\end{minipage}
\caption {(left) AlexNet trained via $\mathrm{StochM3}$ is robust to power distortions.  (right) Test error of our AlexNet for different weight distortions (columns), compared to recent AlexNet models that use binary weights (rows).}
\label{fig:imnet}
\end{figure}

\section{Hypotheses on learning robust networks}
\label{sec:BP-w-proj}

We put forth two ideas on how backprop can find paths to robust solutions.  

First, we explore the idea that imposing constraints on weights can act as a regularizer, similar to DropConnect~\cite{wan2013regularization}.  This idea was first suggested in~\cite{courbariaux2015binaryconnect} for the case of weight binarization.  Here, we examine how imposing weight clipping (without weight projections) can also act as regularizer in the context of proximal methods; see~\cite{parikh2014proximal} for an excellent review on proximal methods.  Consider minimizing $\mathcal{L}(w) + \mathcal{G}(w)$ where $w$ is a vector containing all the weights, $\mathcal{L}(\cdot)$ is the loss, and $\mathcal{G}(\cdot)$ is a regularizer.  For convex $\mathcal{L}(\cdot)$, the proximal gradient method is
$$
w^{t+1} = \mathrm{prox}_\mathcal{G}(w^t - \lambda \frac{\partial \mathcal{L}}{\partial w^t})
$$
where
$$
\mathrm{prox}_\mathcal{G}(v) = \underset{x}{\mathrm{argmin}} \left( \mathcal{G}(x) + (1/2\lambda)||x-v||^2_2 \right).
$$
In the trivial case where $\mathcal{G}(x)=0$, $\mathrm{prox}_\mathcal{G}(v)=v$ and the proximal gradient method reduces to a standard gradient descent.  In our case of weight clipping and assuming a per layer clip value of $1$, $\mathrm{prox}_\mathcal{G}(v)=\mathrm{max}(\mathrm{min}(v, 1),-1)$, where $\mathrm{max}$ and $\mathrm{min}$ functions operate element-wise on $v$.  This corresponds to 
$$
\mathcal{G}(w) = 
	\begin{cases}
		0 & \|w\|_\infty \le 1 \\
		+\infty & \text{otherwise}.
	\end{cases}
$$
In other words, applying weight clipping during an update can be understood (in a convex setting) as imposing a type of regularization that penalizes weights outside the unit ball of the $\ell_\infty$ norm.  Empirically in the non-convex case, this regularization appears to reduces the sensitivity of $\mathcal{L}(w)$ to distortions.  For future work, one could also try other operations on weights commonly used in proximal methods; see~\cite{collins2014memory} for preliminary work in this vein.

Second, consider a stochastic (or deterministic) weight projection.  Let $\phi_{\theta} : \mathbb{R} \rightarrow \mathbb{R}$ be a stochastic {\em projection function} parametrized by a vector $\theta$ that maps a real-valued weight to its projection. We extend $\phi_{\theta}$ to operate on vectors element-wise.  When $w$ is used without projection during forward and backward steps, gradient descent is traversing the cost surface 
$$E_{\text{data}} \left[ \mathcal{L}(w) \right].$$ 
However, if the weights are projected with $\phi_\theta$ before computing forward and backward steps, then backprop only has access to $\mathcal{L}(\phi_{\theta}(w))$.  
By sampling from its distribution over several minibatches, gradient descent is traversing the alternative error surface
$$E_{\text{data}} E_{\phi_{\theta}}[ \mathcal{L}(\phi_{\theta}(w))].$$ 
Hence the solution obtained by minimizing $E_{\text{data}} E_{\phi_{\theta}}[ \mathcal{L}(\phi_{\theta}(w))]$ necessarily provides  the most robustness against distortions under $\phi_{\theta}$ to its weights. Furthermore, the surface $E_{\text{data}} E_{\phi_{\theta}}[ \mathcal{L}(\phi_{\theta}(w))]$ is a smoothed-over version of the surface $E_{\text{data}} \left[ \mathcal{L}(w) \right]$, where the smoothness is controlled by the distribution of the noise source underlying $\phi_{\theta}$'s stochasticity. During training, typically, backprop estimates the gradient of $E_{\text{data}} \left[ \mathcal{L}(w) \right]$ which takes presumed stochasticity of the data into account to produce a smoothed estimate.  In our case, backprop estimates the gradient of $E_{\text{data}} E_{\phi_{\theta}}[ \mathcal{L}(\phi_{\theta}(w))]$ by additionally sampling the weight space in the neighborhood of $w$.  This provides additional gradient smoothing even when $\phi_{\theta}$ is deterministic to some extent. Thus, both the error surface and its gradient are smoother which may explain the empirical results observed in this paper.  During testing, because the objective function is $\mathcal{L}(w)$, using either $w$ directly or with a deterministic projection should yield the best results. For future work, we envision cooling the noise source underlying $\phi_{\theta}$'s stochasticity as training progress so that eventually 
$$E_{\text{data}} E_{\phi_{\theta}}[ \mathcal{L}(\phi_{\theta}(w))] \rightarrow E_{\text{data}} \left[ \mathcal{L}(w) \right]. $$ 
For example, one may draw $\phi_{\theta}(w)$ with respect to the normal distribution ${\cal N}(w, \sigma^2)$ and let $\sigma \rightarrow 0$.  Moreover, we are also interested in extending the domain of $\phi_{\theta}$ to include vectors, so a weight projection can be a function of more than one weight.  For instance, recent work from~\cite{rastegari2016xnor} has already used a projection that depends on the $\ell_1$ norm of the weights for each filter, although their work is in a slightly different context from ours.

Taking a step back, it appears that in all these cases (weight clipping, deterministic projections, and stochastic projections) the weight gradients are distorted relative to the gradients from standard backprop.  Viewing these distortions as a type of noise may bridge our findings with recent work that suggests adding explicit gradient noise results in better overall performance~\cite{neelakantan2015adding}.

\section{Conclusion and future work}
\label{sec:concl}

We expand on previous work demonstrating that networks trained with backprop can become robust to specific weight distortions, such as binary weights.  Here we show that imposing certain weight distortions during training leads to a regime where the network becomes robust not only to that distortion, but to an entire family of distortions as well.  

Based on this observation, we proposed a novel rule that stochastically projects each weight to a random interval based on its current value.  We hypothesize that this rule, similar to the stochastic projection rule in BinaryConnect, is not optimizing the weight values directly, but instead optimizing the neighborhood of the weight vector.  In practice, our rule leads to state of the art performance for CIFAR-10 for both binary and non-binary weighted networks.

Our finding that a network can achieve $89\%$ on CIFAR-10 with $0.68$ bits per weight may also be of practical interest.  One potential application is that weights can be implemented with noisy devices, which could have implications for neuromorphic computing.  

More recently, research on binary weights has been extended to also include binary neuron activations~\cite{courbariaux2016binarynet, rastegari2016xnor, esser2016convolutional}.  Training these networks is similar to the binary weight case, namely binary activations are imposed during training.  We hypothesize that the neuron outputs in these models are also robust to distortions.  If confirmed, this would suggest that imposing other activation constraints could improve performance.  We cannot help but speculate that the built-in robustness to noise in synapses and neurons is an inherent characteristic of the brain and may prove invaluable in opening new directions in deep learning and neuromorphic computing. 



\bibliographystyle{unsrt}
\footnotesize{\bibliography{Ref}}

\begin{thebibliography}{10}

\bibitem{krizhevsky2012imagenet}
Alex Krizhevsky, Ilya Sutskever, and Geoffrey~E Hinton.
\newblock Imagenet classification with deep convolutional neural networks.
\newblock In {\em Advances in neural information processing systems}, pages
  1097--1105, 2012.

\bibitem{silver2016mastering}
David Silver, Aja Huang, Chris~J Maddison, Arthur Guez, Laurent Sifre, George
  Van Den~Driessche, Julian Schrittwieser, Ioannis Antonoglou, Veda
  Panneershelvam, Marc Lanctot, et~al.
\newblock Mastering the game of go with deep neural networks and tree search.
\newblock {\em Nature}, 529(7587):484--489, 2016.

\bibitem{lecun2015deep}
Yann LeCun, Yoshua Bengio, and Geoffrey Hinton.
\newblock Deep learning.
\newblock {\em Nature}, 521(7553):436--444, 2015.

\bibitem{schmidhuber2015deep}
J{\"u}rgen Schmidhuber.
\newblock Deep learning in neural networks: An overview.
\newblock {\em Neural Networks}, 61:85--117, 2015.

\bibitem{johnson2015densecap}
Justin Johnson, Andrej Karpathy, and Li~Fei-Fei.
\newblock Densecap: Fully convolutional localization networks for dense
  captioning.
\newblock {\em arXiv preprint arXiv:1511.07571}, 2015.

\bibitem{han2015learning}
Song Han, Jeff Pool, John Tran, and William Dally.
\newblock Learning both weights and connections for efficient neural network.
\newblock In {\em Advances in Neural Information Processing Systems}, pages
  1135--1143, 2015.

\bibitem{rastegari2016xnor}
Mohammad Rastegari, Vicente Ordonez, Joseph Redmon, and Ali Farhadi.
\newblock Xnor-net: Imagenet classification using binary convolutional neural
  networks.
\newblock {\em arXiv preprint arXiv:1603.05279}, 2016.

\bibitem{merolla2014million}
Paul~A Merolla, John~V Arthur, Rodrigo Alvarez-Icaza, Andrew~S Cassidy, Jun
  Sawada, Filipp Akopyan, Bryan~L Jackson, Nabil Imam, Chen Guo, Yutaka
  Nakamura, et~al.
\newblock A million spiking-neuron integrated circuit with a scalable
  communication network and interface.
\newblock {\em Science}, 345(6197):668--673, 2014.

\bibitem{esser2016convolutional}
Steven~K Esser, Paul~A Merolla, John~V Arthur, Andrew~S Cassidy, Rathinakumar
  Appuswamy, Alexander Andreopoulos, David~J Berg, Jeffrey~L McKinstry, Timothy
  Melano, Davis~R Barch, et~al.
\newblock Convolutional networks for fast, energy-efficient neuromorphic
  computing.
\newblock {\em arXiv preprint arXiv:1603.08270}, 2016.

\bibitem{courbariaux2015binaryconnect}
Matthieu Courbariaux, Yoshua Bengio, and Jean-Pierre David.
\newblock Binaryconnect: Training deep neural networks with binary weights
  during propagations.
\newblock In {\em Advances in Neural Information Processing Systems}, pages
  3105--3113, 2015.

\bibitem{esser2015backpropagation}
Steve~K Esser, Rathinakumar Appuswamy, Paul Merolla, John~V Arthur, and
  Dharmendra~S Modha.
\newblock Backpropagation for energy-efficient neuromorphic computing.
\newblock In {\em Advances in Neural Information Processing Systems}, pages
  1117--1125, 2015.

\bibitem{courbariaux2014training}
Matthieu Courbariaux, Yoshua Bengio, and Jean-Pierre David.
\newblock Training deep neural networks with low precision multiplications.
\newblock {\em arXiv preprint arXiv:1412.7024}, 2014.

\bibitem{stromatias2015robustness}
Evangelos Stromatias, Daniel Neil, Michael Pfeiffer, Francesco Galluppi,
  Steve~B Furber, and Shih-Chii Liu.
\newblock Robustness of spiking deep belief networks to noise and reduced bit
  precision of neuro-inspired hardware platforms.
\newblock {\em Frontiers in neuroscience}, 9, 2015.

\bibitem{soudry2014expectation}
Daniel Soudry, Itay Hubara, and Ron Meir.
\newblock Expectation backpropagation: parameter-free training of multilayer
  neural networks with continuous or discrete weights.
\newblock In {\em Advances in Neural Information Processing Systems}, pages
  963--971, 2014.

\bibitem{glorot2010understanding}
Xavier Glorot and Yoshua Bengio.
\newblock Understanding the difficulty of training deep feedforward neural
  networks.
\newblock In {\em International conference on artificial intelligence and
  statistics}, pages 249--256, 2010.

\bibitem{krizhevsky2009learning}
Alex Krizhevsky and Geoffrey Hinton.
\newblock Learning multiple layers of features from tiny images, 2009.

\bibitem{kingma2014adam}
Diederik Kingma and Jimmy Ba.
\newblock Adam: A method for stochastic optimization.
\newblock {\em arXiv preprint arXiv:1412.6980}, 2014.

\bibitem{ioffe2015batch}
Sergey Ioffe and Christian Szegedy.
\newblock Batch normalization: Accelerating deep network training by reducing
  internal covariate shift.
\newblock {\em arXiv preprint arXiv:1502.03167}, 2015.

\bibitem{abadi2016tensorflow}
Mart{\i}n Abadi, Ashish Agarwal, Paul Barham, Eugene Brevdo, Zhifeng Chen,
  Craig Citro, Greg~S Corrado, Andy Davis, Jeffrey Dean, Matthieu Devin, et~al.
\newblock Tensorflow: Large-scale machine learning on heterogeneous distributed
  systems.
\newblock {\em arXiv preprint arXiv:1603.04467}, 2016.

\bibitem{jia2014caffe}
Yangqing Jia, Evan Shelhamer, Jeff Donahue, Sergey Karayev, Jonathan Long, Ross
  Girshick, Sergio Guadarrama, and Trevor Darrell.
\newblock Caffe: Convolutional architecture for fast feature embedding.
\newblock In {\em Proceedings of the ACM International Conference on
  Multimedia}, pages 675--678. ACM, 2014.

\bibitem{pylearn2_arxiv_2013}
Ian~J. Goodfellow, David Warde-Farley, Pascal Lamblin, Vincent Dumoulin, Mehdi
  Mirza, Razvan Pascanu, James Bergstra, Fr{\'{e}}d{\'{e}}ric Bastien, and
  Yoshua Bengio.
\newblock Pylearn2: a machine learning research library.
\newblock {\em arXiv preprint arXiv:1308.4214}, 2013.

\bibitem{lin2013network}
Min Lin, Qiang Chen, and Shuicheng Yan.
\newblock Network in network.
\newblock {\em arXiv preprint arXiv:1312.4400}, 2013.

\bibitem{moerland1997neural}
Perry Moerland and Emile Fiesler.
\newblock Neural network adaptations to hardware implementations.
\newblock Technical report, IDIAP, 1997.

\bibitem{ILSVRC15}
Olga Russakovsky, Jia Deng, Hao Su, Jonathan Krause, Sanjeev Satheesh, Sean Ma,
  Zhiheng Huang, Andrej Karpathy, Aditya Khosla, Michael Bernstein,
  Alexander~C. Berg, and Li~Fei-Fei.
\newblock {ImageNet Large Scale Visual Recognition Challenge}.
\newblock {\em International Journal of Computer Vision (IJCV)},
  115(3):211--252, 2015.

\bibitem{vedaldi2015matconvnet}
Andrea Vedaldi and Karel Lenc.
\newblock Matconvnet: Convolutional neural networks for matlab.
\newblock In {\em Proceedings of the 23rd Annual ACM Conference on Multimedia
  Conference}, pages 689--692. ACM, 2015.

\bibitem{wan2013regularization}
Li~Wan, Matthew Zeiler, Sixin Zhang, Yann~L Cun, and Rob Fergus.
\newblock Regularization of neural networks using dropconnect.
\newblock In {\em Proceedings of the 30th International Conference on Machine
  Learning (ICML-13)}, pages 1058--1066, 2013.

\bibitem{parikh2014proximal}
Neal Parikh and Stephen~P Boyd.
\newblock Proximal algorithms.
\newblock {\em Foundations and Trends in optimization}, 1(3):127--239, 2014.

\bibitem{collins2014memory}
Maxwell~D Collins and Pushmeet Kohli.
\newblock Memory bounded deep convolutional networks.
\newblock {\em arXiv preprint arXiv:1412.1442}, 2014.

\bibitem{neelakantan2015adding}
Arvind Neelakantan, Luke Vilnis, Quoc~V Le, Ilya Sutskever, Lukasz Kaiser,
  Karol Kurach, and James Martens.
\newblock Adding gradient noise improves learning for very deep networks.
\newblock {\em arXiv preprint arXiv:1511.06807}, 2015.

\bibitem{courbariaux2016binarynet}
Matthieu Courbariaux and Yoshua Bengio.
\newblock Binarynet: Training deep neural networks with weights and activations
  constrained to+ 1 or-1.
\newblock {\em arXiv preprint arXiv:1602.02830}, 2016.

\end{thebibliography}

\end{document}